\pdfoutput=1
\documentclass[letterpaper, 10 pt, conference]{ieeeconf}

\IEEEoverridecommandlockouts
\usepackage{newtxtext}
\usepackage{newtxmath}
\usepackage{cite}
\usepackage{graphicx}
\usepackage[caption=false,font=footnotesize]{subfig}
\usepackage{diagbox}
\usepackage{array}
\usepackage{color}
\usepackage{units}
\usepackage{url}

\usepackage{verbatim}
\usepackage{textcomp}
\usepackage{bm}
\usepackage{bbm}
\usepackage{mathtools}
\usepackage{cleveref}
\usepackage{amsmath} 

\usepackage{amssymb}

\newtheorem{remark}{Remark}

\usepackage{algpseudocode}
\usepackage[linesnumbered,ruled,vlined]{algorithm2e}

\DeclareMathOperator*{\argmin}{arg\,min}

\newtheorem{problem}{Problem}
\newtheorem{proposition}{Proposition}
\SetKwInput{KwInput}{Input}                
\SetKwInput{KwOutput}{Output}              

\title{\LARGE \bf
MDCPP: Multi-Robot Dynamic Coverage Path Planning \\ for Workload Adaptation}

\author{Jun Chen, Mingjia Chen, Tianlong Yu, Qi Nie, and Shinkyu Park%
  \thanks{J.~Chen, M.~Chen, T.~Yu, and Q.~Nie are with the School of Electrical and Automation Engineering, Nanjing Normal University, Nanjing, China.}%
  \thanks{S.~Park is with the Computer, Electrical and Mathematical Sciences and Engineering Division, King Abdullah University of Science and Technology, Thuwal, Saudi Arabia.}}

\begin{document}

\maketitle

\begin{abstract}
Multi-robot coverage path planning commonly balances geometric area or path length under a constant-speed assumption. This assumption is inadequate when sensing or interaction tasks cause spatially varying traversal speeds, because equal areas can induce markedly different completion times. We propose Multi-Robot Dynamic Coverage Path Planning (MDCPP), which learns a Gaussian-mixture workload field from partial observations, predicts cell-wise service times, and repeatedly repartitions the uncovered cells through a distributed capacity-constrained assignment. We establish finite termination and pairwise local optimality of each synchronized assignment round, bound the service-time makespan degradation due to estimation error, and state sufficient conditions for complete coverage. A 600-run benchmark against sweeping, LS-MCPP, reactive reassignment, and an oracle shows that prediction is most valuable under strong heterogeneity and improves aggregate paired makespan over the nonpredictive alternatives. A three-UGV experiment further validates route execution and spatial speed adaptation under localization, drivetrain, and wireless-control effects.
\end{abstract}


\section{Introduction}
Coverage path planning (CPP) generates paths that collectively observe or service a region \cite{galceran2013survey}. Classical methods optimize distance, time, or energy \cite{oksanen2009coverage,cabreira2018energy,otto2018optimization}, with variants for building-constrained, underwater, or boustrophedon team coverage \cite{chen2024improved,zhu2019complete,rekleitis2008efficient}. Multi-robot CPP (MCPP) additionally partitions the workspace: cellular and spanning-tree methods balance geometric workload \cite{karapetyan2017efficient,tang2021mstc}, scalable methods handle non-convex regions \cite{collins2021scalable}, and learning or distributed methods improve graph-based and unknown-space coverage \cite{tolstaya2021multi,hu2023multi}. For known geometry or costs, recent methods use weighted-terrain mixed-integer optimization \cite{tang2023mip}; LS-MCPP combines Extended-STC, local search, and MAPF post-processing for scalable path deconfliction \cite{tang2025lsmcpp}. Dynamic spanning-tree, on-demand/concurrent, APF, and reconfigurable methods adapt paths or ownership online \cite{mo2024mdstc,mitra2024ondemand,mitra2025concurrent,wang2024apf,luo2026reconfigurable,kim2026graph}. MAC-Planner couples coarse-map task allocation with online planning and updates assignments from real-time completion status \cite{wang2025mac}. These methods adapt geometry, ownership, or known costs; MDCPP instead infers an unknown service-time field and partitions by predicted remaining workload.

Unknown-environment CPP incrementally constructs decompositions as free space is revealed \cite{acar2002sensor}, while cooperative methods address communication and team robustness \cite{fazli2010complete}. Those settings principally learn traversability or coordinate geometric coverage. Here, the map geometry is known but the spatially varying service demand is not: observations made while covering a cell change the predicted time cost of every unvisited cell. This distinction couples online estimation, heterogeneous robot speed, and repeated allocation in a way that a purely geometric frontier or fixed partition does not capture.

These methods do not directly address missions in which service cost is spatially structured, initially unknown, and revealed during coverage. In coral imaging, pollution sampling, or crop inspection, robots transit quickly through low-demand regions but slow down or dwell near targets. Equal-area or fixed-edge-cost partitions can therefore overload a slow robot or one assigned a high-density region. Multi-Robot Dynamic Coverage Path Planning (MDCPP) learns this latent workload field and repartitions uncovered cells by predicted remaining completion time. The main contributions are:
\begin{itemize}
    \item a workload-aware MCPP formulation that explicitly combines travel and spatially varying service time;
    \item an online GMM estimator and a distributed, event-triggered repartitioning method, together with termination, local-optimality, completeness, and estimation-robustness results; and
    \item a 600-run comparison with geometric, reactive, and oracle baselines, plus a three-UGV validation of path execution and Gaussian speed modulation.
\end{itemize}

\section{Problem Formulation}

Consider coral-reef monitoring by UUVs. A vehicle transits quickly through inspected water but must slow down to acquire high-resolution imagery and perform local analysis near dense coral populations. Without a prior density map, equal-area assignment can give one UUV most sensing-intensive cells. MDCPP uses observations from serviced cells to predict remaining demand and redistribute monitoring tasks before a bottleneck is fully exposed.

To formalize, we denote a team of $n_r$ robots as $R = \{r_1, \ldots, r_{n_r}\}$. 
Each robot is equipped with an isotropic sensor with sensing range $l_{\text{sensing}}$ and has self-localization capability. 
Anisotropic detection models may be mapped to equivalent isotropic models \cite{chen2021distributed}. 
Each robot $r_i$ can move at a maximum speed $v_{\max,i}$.
Its coverage speed $v_{\mathrm{cov},i}$, however, varies with task-specific sensing or interaction requirements and environmental conditions.
Robots exchange poses, detections, and assignments with communication neighbors at synchronized reassignment rounds.

Consider a known convex task space $E_0 \subset \mathbb R^2$ containing stationary targets at unknown locations $T\subset E_0$.
The distribution of targets is modeled using Gaussian mixtures, which represent spatial aggregation caused by diffusion, biological behavior, or local sources \cite{crawford2020use}. 
We model the target distribution as a mixture of $K$ Gaussian components, each centered at $\mu_k (\mu_{k,x}, \mu_{k,y})$ with a standard deviation $\sigma_k$, where $k\in \{ 1,2,\ldots,K \}$. 

At each decision epoch $t$, let $E_t$ denote the uncovered part of $E_0$. We discretize it into $n_{g,t}$ uniform square cells $G_t = \{g_1, \ldots, g_{n_{g,t}}\}$ whose union is $E_t$. 
We select the cell size from the sensor footprint so that one prescribed service action covers a cell.
By slight abuse of notation, we use $g_1, g_2, \ldots$ to denote the centroids of the corresponding squares.
The higher the mixture value within a grid cell, the greater the target density in that cell.
Therefore, when interacting with targets in such a grid cell, the robot must adopt a lower coverage velocity.

Passing through an uncovered cell at travel speed does not service it; the cell leaves $G_t$ only after sensing or interaction is complete. Covered cells may subsequently be crossed at $v_{\max,i}$.

Our goal is to assign all uncovered grid cells to the robots, forming a partition into $n_r$ disjoint subsets $S_i\subseteq G_t$ such that $\bigcup_{i=1}^{n_r}S_i=G_t$ and $S_i\cap S_{i'}=\emptyset$ for all $i\neq i'$.
For each robot $r_i$, the predicted completion time of an assigned partition $S_i$ comprises the travel time along a path visiting its cells and the cell-wise service time. Specifically,
\begin{equation}
\widehat T_i(S_i,\mathbf p_i) = \frac{L(\mathbf p_i)}{v_{\max,i}} + \sum_{g\in S_i}\widehat c_{ig},
\label{eq:predicted_completion}
\end{equation}
where $L(\mathbf p_i)$ is the inter-cell transit length and $\widehat c_{ig}$ is the service-time estimate for cell $g$, including its sensing or interaction action. Thus, transit and service are not double-counted.
\begin{problem}[Cooperative coverage optimization]
At each decision epoch, the objective is to minimize the predicted makespan,
\begin{equation}
\min_{\{S_i,\mathbf p_i\}_{i=1}^{n_r}} \max_{i=1,\dots,n_r} \widehat T_i(S_i,\mathbf p_i),   
\end{equation}
by jointly choosing a partition and a route for every robot. MDCPP approximates this coupled problem through three modules: (i) online estimation of cell service times from partial observations, (ii) cell assignment that reduces the largest predicted remaining workload, and (iii) route ordering that reduces transit distance within each assigned set.
\label{prob:optimize}
\end{problem}

The assignment module accounts for both spatial workload and heterogeneous robot capability.
At time $t$, we use an additive surrogate for the remaining workload of robot $r_i$:
\begin{equation}
    \widehat w_{i,t}(S_{i,t}) = \widehat \tau_i(q_{i,t},S_{i,t})
    + \sum_{g\in S_{i,t}} \widehat c_{ig,t},
\label{eq:workload}
\end{equation}
where $\widehat\tau_i$ is the transit time from the current pose through a candidate cell order and $\widehat c_{ig,t}$ is the robot-specific service-time prediction obtained from the learned workload field and robot $i$'s speed limits. The assignment routine minimizes the maximum of this surrogate; the route is subsequently refined by the path planner.

\begin{remark}
For simplicity, $E_0$ is convex and transit cost is the Euclidean distance from the robot's current position to a goal. In an obstructed workspace, the same algorithms can instead use the length of a feasible collision-free path.
\label{remark:euclidean}
\end{remark}

Accordingly, the assignment module solves the following problem.
\begin{problem}[Optimized Cell Assignment] \label{problem:workload_assignment}
\begin{align}
\min_{\{S_{i,t}\}_{i=1}^{n_r}} \quad &
\max_{i \in \{1, \ldots, n_r\}} \widehat w_{i,t}(S_{i,t})
\label{eq:minimize}
\\[-1mm]
\mathrm{s.t.}\quad &\bigcup_i S_{i,t}=G_t,\qquad S_{i,t}\cap S_{j,t}=\varnothing\quad(i\ne j).
\end{align}
\end{problem}

\section{Dynamic Coverage Path Planning}

\subsection{Estimating Target Distribution}
The latent target density at a cell centered at $(x,y)$ is modeled by a bounded sum of $K$ isotropic Gaussian components:
\begin{equation}
\rho_{\mathrm{true}}(x,y)=\operatorname{clip}_{[0,1]}\!\left[
\sum_{k=1}^{K}A_k\exp\!\left(-\frac{\|(x,y)-\mu_k\|^2}{2\sigma_k^2}\right)\right],
\label{eq:rho_true}
\end{equation}
where $A_k\in(0,1]$ is the amplitude of component $k$, $\mu_k=(\mu_{k,x},\mu_{k,y})$ is its center, and $\sigma_k$ is its standard deviation. The illustrative simulation uses unit amplitudes; the Monte Carlo benchmark samples $A_k\sim\mathcal U[0.75,1]$ and clips the component sum as shown in \eqref{eq:rho_true}. Thus, $A_k$ is a component amplitude and need not equal the peak of the clipped mixture when components overlap. Robot $r_i$ observes
\begin{equation}
z_i(x,y)=\rho_{\mathrm{true}}(x,y)+\epsilon_i,\qquad
\epsilon_i\sim\mathcal N(0,\sigma_{\mathrm{noise},i}^2),
\label{eq:z}
\end{equation}
where the noise variance can depend on the robot. From the observations $\mathcal O_t=\{(g_j,z_j)\}$ collected by the team, the estimator selects the component count and estimates centers, spreads, amplitudes, and a nonnegative background term.

\subsubsection{Candidate Centers}
Coordinates whose observations exceed a threshold $\theta$ form $Z_t=\{g_j:(g_j,z_j)\in\mathcal O_t,\ z_j>\theta\}$. For each candidate $K$, weighted K-means partitions $Z_t$ by solving
\begin{equation}
\min_{\{W_k,\mu_k\}}\sum_{k=1}^{K}\sum_{g_j\in W_k}z_j\|g_j-\mu_k\|^2.
\end{equation}
The weighted cluster centroids provide candidate centers. Restricting center estimation to sufficiently large observations prevents low-density background cells from dominating the fit.

\subsubsection{Model Selection and Parameter Fitting}
For each set of candidate centers and spreads, least squares followed by projection onto the nonnegative orthant estimates a background $b$ and amplitudes $A_k$. Model order and spreads minimize a complexity-penalized fit:
\begin{equation}
\frac{1}{|\mathcal O_t|}\sum_{(g_j,z_j)\in\mathcal O_t}
\left[z_j-b-\sum_{k=1}^{K}A_k e^{-\|g_j-\mu_k\|^2/(2\sigma_k^2)}\right]^2+\lambda K.
\label{eq:gmm_fit}
\end{equation}
The Monte Carlo implementation uses $\theta=0.55$, $K\leq3$, a shared spread grid $\{2,2.5,\ldots,4.5\}$ cells, and $\lambda=0.0025$. It requires three thresholded samples per component; with fewer than eight observations or four above $\theta$, it uses the observed mean as a constant prior (zero initially). The resulting field is
\begin{equation}
\widehat\rho_t(x,y)=\operatorname{clip}_{[0,1]}\!\left[
\hat b+\sum_{k=1}^{\widehat K}\widehat A_k
e^{-\|(x,y)-\widehat\mu_k\|^2/(2\widehat\sigma_k^2)}\right].
\end{equation}
Measurements overwrite predictions at observed cells. Interpolating speed between each robot's travel and detection limits using $\widehat\rho_t(g)$ gives, for equivalent service length $\ell_g$ (the cell side in the benchmark),
\begin{equation}
\widehat c_{ig}=\frac{\ell_g}{v_{\mathrm{tr},i}-(v_{\mathrm{tr},i}-v_{\mathrm{det},i})\widehat\rho_t(g)}.
\label{eq:service_map}
\end{equation}
Because $0\leq\widehat\rho_t(g)\leq1$ and $0<v_{\mathrm{det},i}\leq v_{\mathrm{tr},i}$, \eqref{eq:service_map} is finite, monotone in predicted density, and bounded between $\ell_g/v_{\mathrm{tr},i}$ and $\ell_g/v_{\mathrm{det},i}$. The assignment therefore compares quantities in seconds rather than mixing density, distance, and time. Measurements in \eqref{eq:z} are clipped to the admissible interval before this conversion.

The estimator deliberately remains small: the candidate center count and spread grid are bounded independently of workspace size. With $N_t=|\mathcal O_t|$, at most $K_{\max}$ components, and $N_\sigma$ spread candidates, the least-squares stage costs $O\!\left(N_\sigma\sum_{K=1}^{K_{\max}}(N_tK^2+K^3)\right)$, in addition to weighted K-means. Because $K_{\max}=3$ here, this stage scales linearly with the number of observations. This bounded model class is suitable for localized hotspots but is not intended to represent arbitrary multimodal or discontinuous fields.

\subsection{Performance Properties}
The following results clarify what the online assignment guarantees. Define the round potential as $\Phi_t(\mathcal S)=\max_i\widehat w_{i,t}(S_{i,t})$. Let $c_{ig}$ be the true cell service time and assume that a synchronized reassignment round accepts an exchange only if it strictly decreases $\Phi_t$.

\begin{proposition}[Termination and local optimality]
For a finite uncovered grid and a fixed communication graph, each synchronized reassignment round terminates after finitely many accepted exchanges. At termination, no feasible exchange between a communicating pair strictly decreases $\Phi_t$.
\end{proposition}
\emph{Proof.} There are finitely many assignments of $n_{g,t}$ labeled cells to $n_r$ robots. Every accepted exchange strictly decreases the potential, so an assignment cannot recur. The procedure therefore terminates in finitely many exchanges. The stopping test gives pairwise local optimality on the communication graph. \hfill$\square$

This is a local, rather than global, guarantee; graph connectivity alone does not imply equality with a centralized optimum.

\begin{proposition}[Estimation robustness]
Suppose $|\widehat c_{ig}-c_{ig}|\leq\varepsilon$ for every remaining robot-cell pair and every admissible partition assigns at most $M$ cells to one robot. If $\widehat{\mathcal S}$ minimizes the estimated service-time makespan and $\mathcal S^\star$ minimizes the true one, then
\begin{equation}
\max_i C_i(\widehat S_i)\leq \max_i C_i(S_i^\star)+2M\varepsilon,
\label{eq:robust_bound}
\end{equation}
where $C_i(S)=\sum_{g\in S}c_{ig}$.
\end{proposition}
\emph{Proof.} For any $S_i$, $|\widehat C_i(S_i)-C_i(S_i)|\leq |S_i|\varepsilon\leq M\varepsilon$. Thus the true makespan of the estimated optimum is at most its estimated makespan plus $M\varepsilon$, which is no greater than the estimated makespan of $\mathcal S^\star$ plus $M\varepsilon$. Applying the same bound to $\mathcal S^\star$ proves \eqref{eq:robust_bound}. \hfill$\square$

This bound isolates estimation error for the ideal assignment objective; the distributed exchange routine additionally incurs its pairwise-local optimization gap.
It also applies only to the service component of \eqref{eq:predicted_completion}. If the transit-time estimate has a uniform error bound $\delta$ per robot, the same argument adds $2\delta$ to the right-hand side of \eqref{eq:robust_bound}. Consequently, estimation accuracy and route approximation affect the bound through separate additive terms.

The density-to-time conversion makes this result directly interpretable. For $f_i(\rho)$ defined by \eqref{eq:service_map},
\begin{equation}
|f_i'(\rho)|\leq L_i:=\frac{\ell_g(v_{\mathrm{tr},i}-v_{\mathrm{det},i})}{v_{\mathrm{det},i}^2}.
\label{eq:density_lipschitz}
\end{equation}
Thus $|\widehat\rho_t(g)-\rho_{\mathrm{true}}(g)|\leq\eta$ implies $|\widehat c_{ig}-c_{ig}|\leq L_i\eta$, so Proposition~2 uses $\varepsilon=\max_iL_i\eta$. Since $L_i$ grows rapidly as $v_{\mathrm{det},i}$ decreases, the same density error causes a larger time-cost error for a slow-service robot. This is a conditional sensitivity result, not a claim that sparse observations always identify the GMM; receding-horizon updates instead overwrite predictions at measured cells and repeatedly refit the remaining field.

\subsection{Optimized Cell Assignment}
MDCPP has three assignment stages. First, weighted Lloyd iterations spread the robots toward initial goals. Second, the robots dynamically partition $G_t$ subject to workload capacity. Third, each robot orders the cells in its assigned region.

\subsubsection{Determine Initial Goals}

The robots first spread across the area by constructing a power diagram and moving toward weighted cell centroids using Lloyd iterations \cite{cortes2004coverage}.
Let $\mathcal S=\{S_1,\ldots,S_{n_r}\}$, $Q=\{q_1,\ldots,q_{n_r}\}$, and let $\beta_i$ be a power weight reflecting robot $i$'s predicted capacity. The initialization minimizes
\begin{equation}
\mathcal H(Q,\mathcal S)=\sum_{i=1}^{n_r}\sum_{g_j\in S_i}
\left(\|g_j-q_i\|^2-\beta_i^2\right),
\label{eq:optimization}
\end{equation}
over the partition and robot positions. Minimization over $\mathcal S$ induces a power partition, with $\beta_i$ adjusted so that cell loads reflect predicted workload and robot capability. Minimization over $Q$ moves each robot toward the centroid of its power cell.
Iterations stop when all centroid displacements fall below $\epsilon_s$. This spreads the team toward positions that reduce \eqref{eq:optimization} before online coverage begins.

\subsubsection{Cell Assignment}

\begin{algorithm}[tbp]
\small
\DontPrintSemicolon
\For{each ID-ordered communicating pair $(i,j)$}{
Exchange poses, assignments, and predicted cell costs\;
Build priority queues $H_i,H_j$ of feasible cell transfers\;
\While{a queued transfer strictly decreases the potential}{
Commit the best move preserving a disjoint complete partition\;
Update the two workloads and affected queue keys\;
}
Send the agreed ownership map to the paired robot\;
}
\Return $\mathcal S_{t+1}$\;
\caption{Distributed cell assignment} 
\label{alg:distributed}
\end{algorithm}

Algorithm~\ref{alg:distributed} extends capacity-constrained Voronoi exchange \cite{balzer2009capacity} to predicted remaining time. Unique robot IDs serialize communicating pairs. Candidate cell transfers are kept in priority queues; applications that require connected territories may restrict candidates to ownership boundaries. A geometric queue key is
\begin{equation}
\Delta e(g,q_i,q_j)=\|g-q_i\|-\|g-q_j\|,
\label{eq:key}
\end{equation}
where positive $\Delta e$ favors assigning $g$ to $r_j$. The final decision evaluates the resulting workload and partition feasibility, so geometric proximity alone cannot approve a move.

Under unlimited communication, every robot pair is eligible for an exchange, and the terminal partition is pairwise locally optimal over the complete graph. Under range constraints, the same statement holds only for edges present in the communication graph during that synchronized round. The method is therefore not claimed to recover a globally optimal capacity-constrained partition.
Lower-ID robots may evaluate more pairs, so IDs can be assigned according to onboard computation or rotated between rounds.

Each pairwise negotiation exchanges only poses, relevant ownership records, and predicted costs for candidate cells; the full raw observation history need not be transmitted. Queue entries touching an accepted cell are invalidated and recomputed before the next candidate is tested. This preserves partition consistency while avoiding a centralized assignment solve. The price is conservatism: a transfer that would be beneficial only as part of a multi-robot cycle may not be accepted by any individual pair. Periodic ID rotation can reduce persistent ordering bias but does not alter the graph-local nature of Proposition~1.

If a communicating pair considers $b_{ij}$ candidate cells, one negotiation transmits $O(b_{ij})$ ownership and cost records and stores $O(b_{ij})$ queue entries. A heap gives $O(\log b_{ij})$ key updates after an accepted move. The finite-assignment proof supplies a worst-case termination guarantee but not a useful polynomial bound on the number of moves; in practice, strict makespan decrease and event-triggered invocation limit repeated negotiations. Synchronization is required only within a round: stale or missing neighbors defer their exchanges rather than invalidating the ownership partition.

\subsection{Coverage Path Planning}

Given $S_i$, robot $r_i$ seeks an open route that starts at its current position $q_i$ and visits every assigned cell once.
\begin{problem}[Shortest coverage route]\label{problem:path}
\begin{equation}
\begin{aligned}
\mathbf p_i^\star\in\argmin_{\mathbf p_i\in\operatorname{Perm}(S_i)}\ \ &
\operatorname{dist}(q_i,p_i(1))\\[-1mm]
&+\sum_{k=1}^{|S_i|-1}\operatorname{dist}(p_i(k),p_i(k+1)).
\end{aligned}
\label{eq:shortest_path}
\end{equation}
\end{problem}
Here, $\operatorname{Perm}(S_i)$ denotes all orderings of $S_i$. Euclidean distance is used in the convex obstacle-free setting; a collision-free shortest-path metric can replace it in an obstructed environment. This is an open Hamiltonian-path problem. We use the nearest-neighbor heuristic \cite{applegate2011traveling} for computational efficiency.

\begin{proposition}[Complete coverage]
Assume that (i) the finite uncovered cells remain partitioned without omission or duplication, (ii) each robot remains operational and triggered rounds finish without indefinitely preempting execution, (iii) the controller reaches and services its next waypoint in finite time, and (iv) replanning never reintroduces a covered cell. Then MDCPP covers every cell in finite time.
\end{proposition}
\emph{Proof.} At any decision epoch, each uncovered cell belongs to exactly one finite route. By (iii), at least one route advances to a new cell in finite time. That cell is removed permanently by (iv), so $|G_t|$ decreases monotonically. Since $|G_0|$ is finite, after at most $|G_0|$ such coverage events the uncovered set is empty. \hfill$\square$

A robot requests repartitioning when $|S_{i,t}|<n_0$; a larger $n_0$ triggers earlier redistribution at the cost of more communication and computation. Algorithm~\ref{alg:mdcpp} summarizes execution: robots plan routes over their initial cells, adapt speed while servicing them, and repartition and replan when any neighbor is close to exhausting its route.

\begin{algorithm}[tbp]
\small
\DontPrintSemicolon
Initialize goals by weighted Lloyd iterations (offline) \;
Initialize cell assignment using Algorithm \ref{alg:distributed} (offline) \;
\For{\rm each robot $r_i$ in parallel}{
Move to the initial goal at $v_{\max,i}$ \;
Plan a coverage path $P_i$ for Problem~\ref{problem:path} \;
\While{\rm $G_t\neq\varnothing$}{
\If{\rm Repartition requested by $r_j \in \mathcal{N}_i(t)$ or $|S_{i,t}|<n_0$}{
Update $S_{i,t}$ using Algorithm \ref{alg:distributed} \;
Plan a coverage path $P_i$ for Problem~\ref{problem:path} \;
}
\If{\rm $q_i$ is within cell $P_i(k)$}{
Set $v_i$ from the task-specific service law and $\widehat\rho_t(P_i(k))$ \;
}
\Else{
$v_i = v_{\max,i}$ \;
}
Move along $P_i$ with speed $v_i$\;
\If{service of $P_i(k)$ is complete}{record its observation and remove $P_i(k)$ from $G_t$\;}
}
}
\caption{MDCPP} 
\label{alg:mdcpp}
\end{algorithm}

\section{Simulations}
We evaluate MDCPP in a $20\times20$ grid of $10\,\mathrm{m}\times10\,\mathrm{m}$ cells using four robots with a $7.07\,\mathrm{m}$ isotropic sensing radius and $n_0=2$.

\subsection{Dynamic Cell Assignment}
\begin{figure}[!t]
    \centering
    \subfloat[Target field]{\includegraphics[width=0.38\columnwidth]{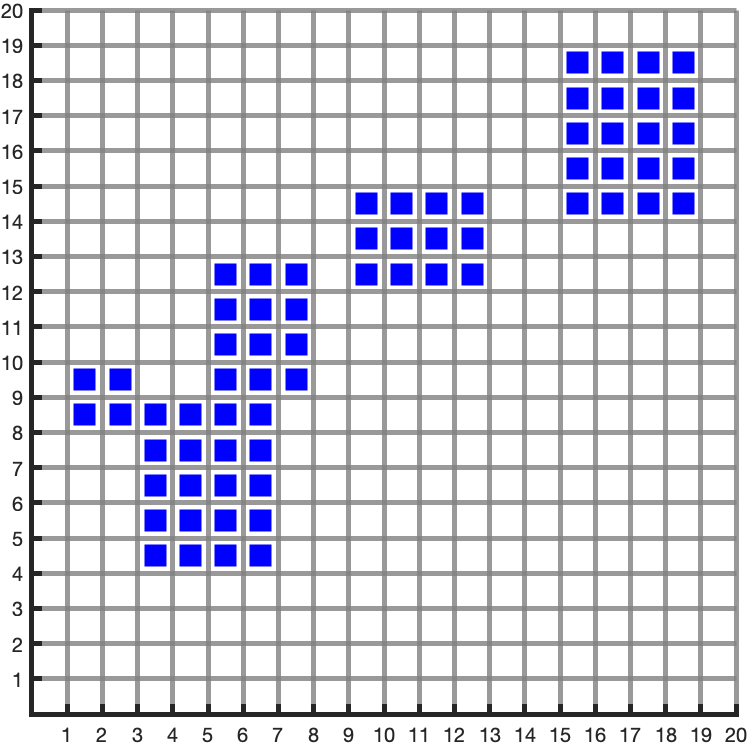}}\hfill
    \subfloat[Static sweep]{\includegraphics[width=0.38\columnwidth]{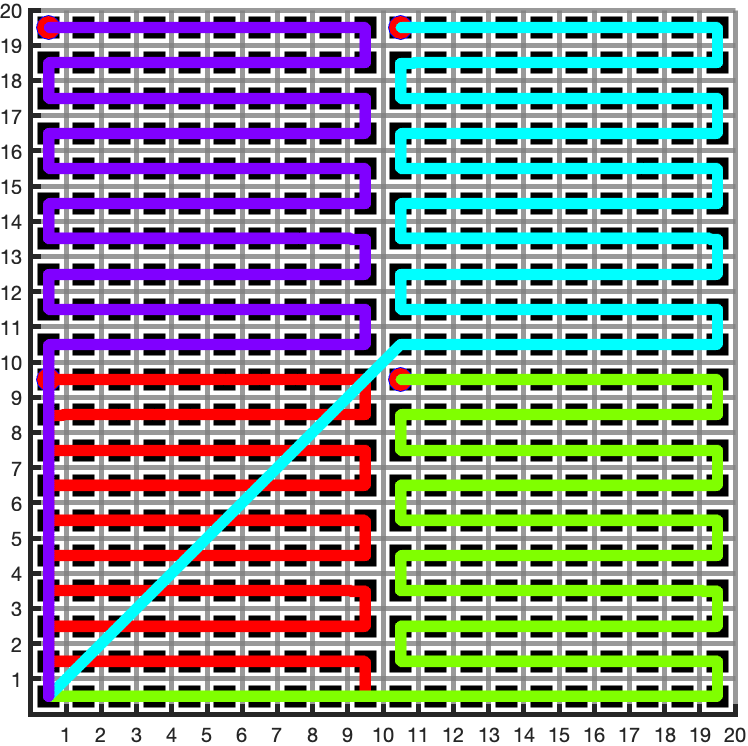}}
    \caption{Simulation setup: (a) a $20\times20$ grid with targets in blue cells; (b) four fixed rectangular sweeps.}
    \label{fig:setup}
\end{figure}

For Problem~\ref{problem:workload_assignment}, targets occupy the blue cells in Fig.~\ref{fig:setup}(a), while Fig.~\ref{fig:setup}(b) shows the fixed rectangular-sweep reference. In this illustrative simulation, the instantaneous speeds in the travel, detection, and interaction modes vary independently within $50\%$--$150\%$ of their nominal values. The corresponding nominal triplets for robots 1--4 are $3.0/1.0/0.10$, $2.0/0.8/0.12$, $5.0/0.6/0.08$, and $4.0/0.12/0.06\,\mathrm{m/s}$.

\begin{figure*}[t]
    \centering
    \subfloat[75 s]{\includegraphics[width=0.155\textwidth]{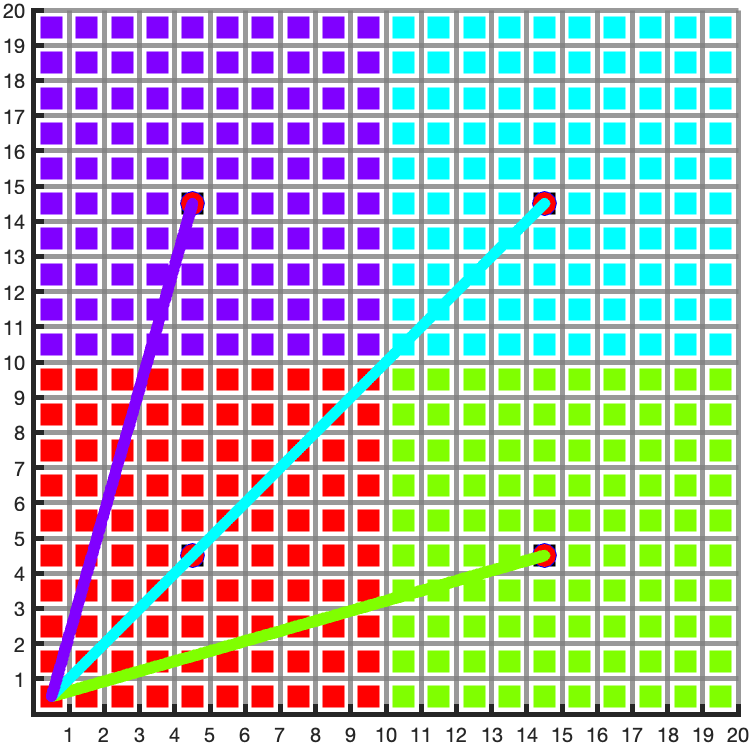}\label{fig:s1}}\hfill
    \subfloat[1294 s]{\includegraphics[width=0.155\textwidth]{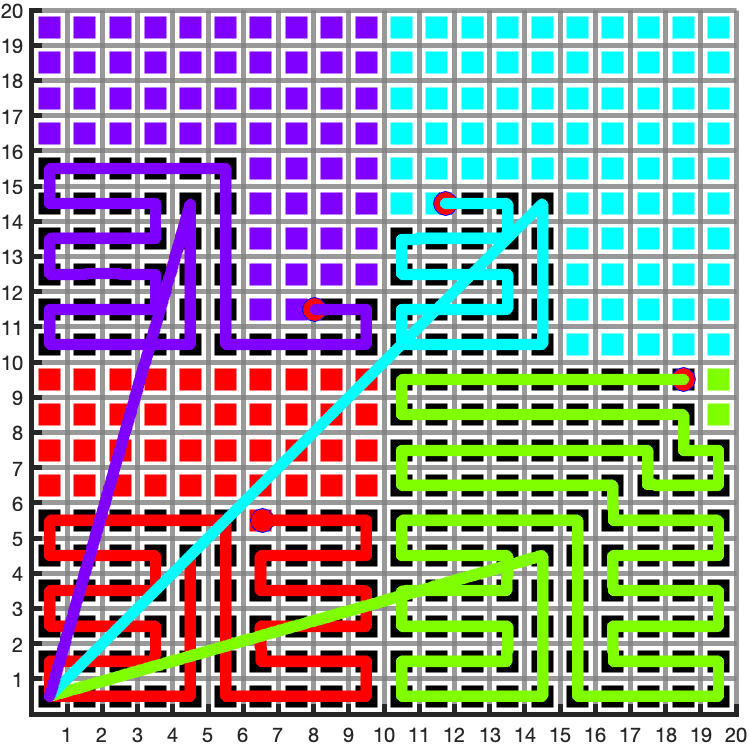}\label{fig:s2}}\hfill
    \subfloat[1295 s]{\includegraphics[width=0.155\textwidth]{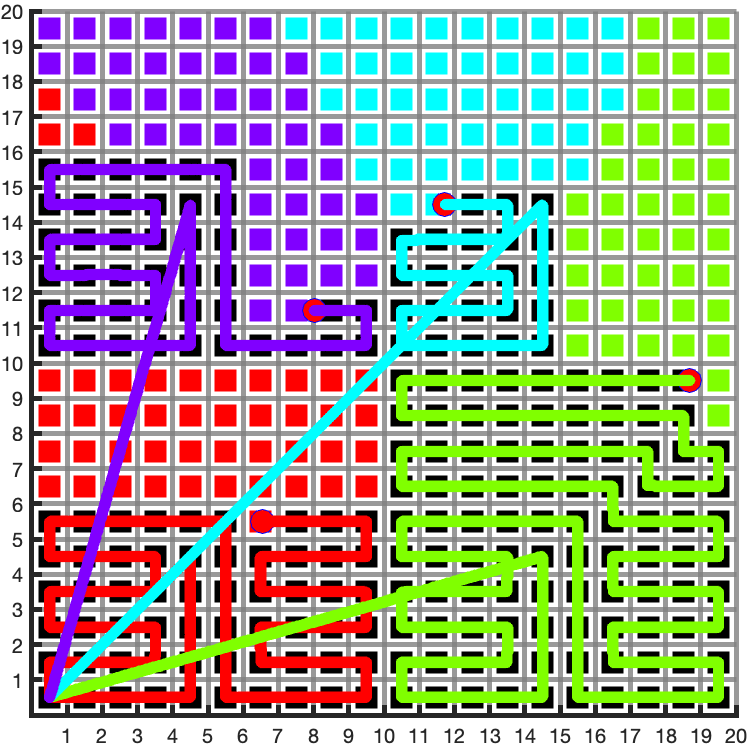}\label{fig:s3}}\hfill
    \subfloat[2691 s]{\includegraphics[width=0.155\textwidth]{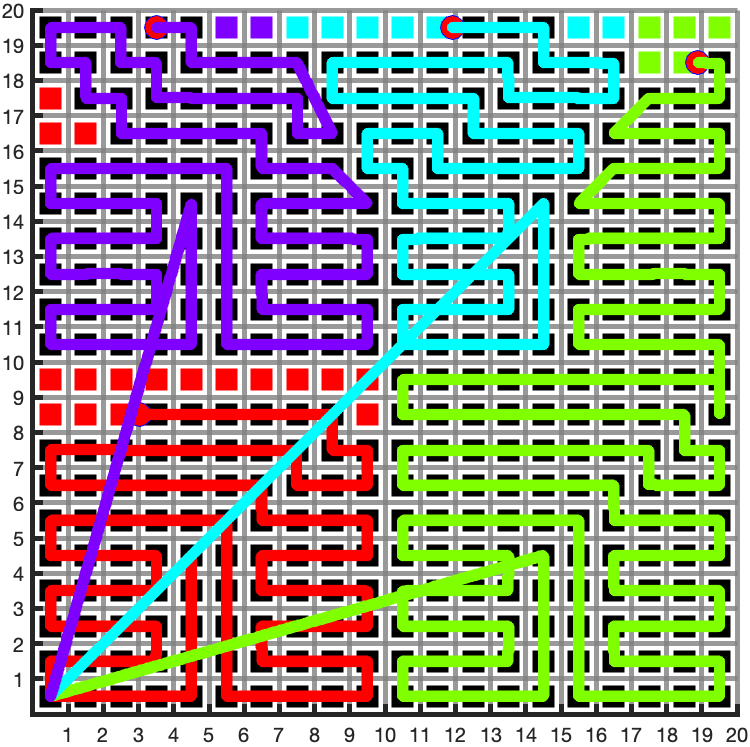}\label{fig:s4}}\hfill
    \subfloat[2692 s]{\includegraphics[width=0.155\textwidth]{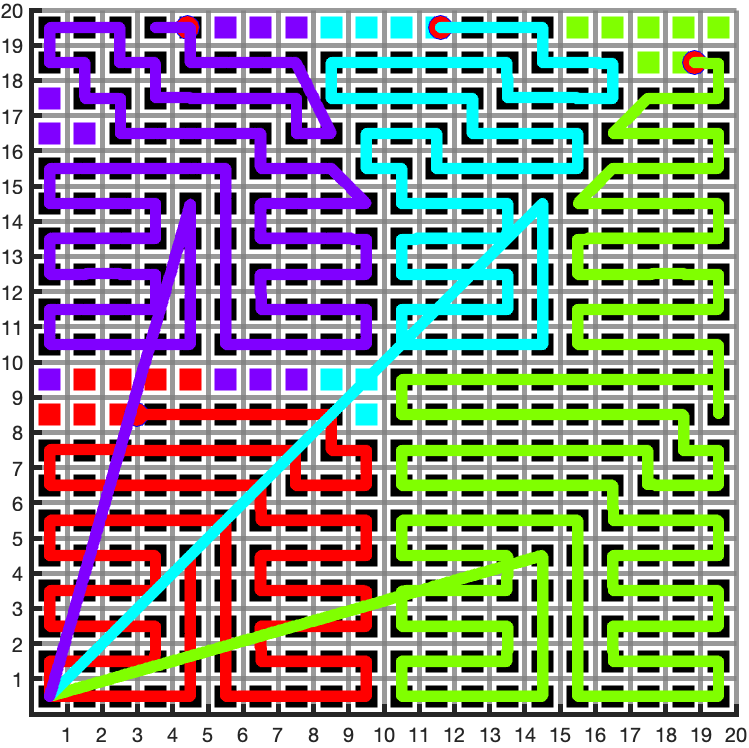}\label{fig:s5}}\hfill
    \subfloat[3174 s]{\includegraphics[width=0.155\textwidth]{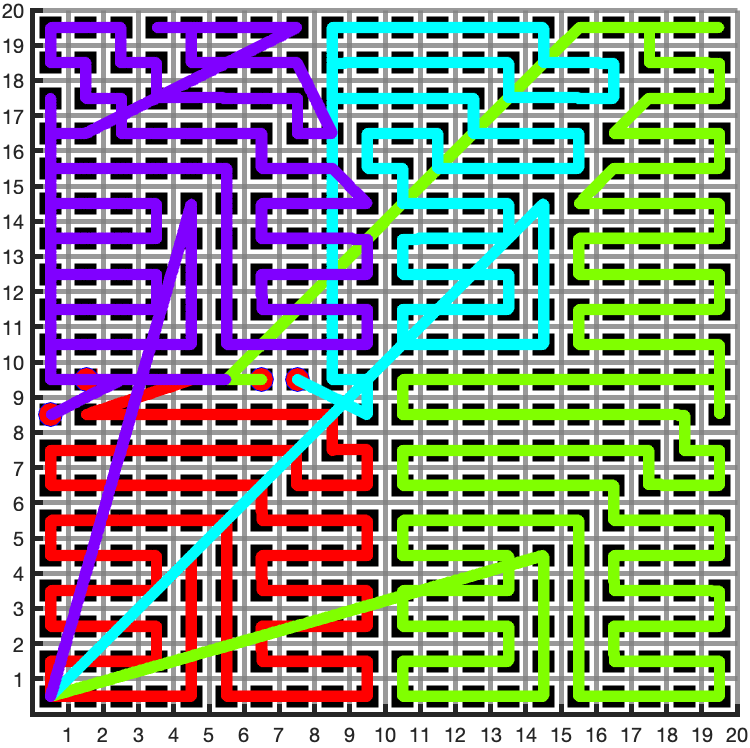}\label{fig:s6}}
    \caption{Dynamic coverage sequence. The consecutive pairs (b)--(c) and (d)--(e) show workload-triggered reassignment of uncovered cells.}
    \label{fig:reassignment10}
\end{figure*}

Figure~\ref{fig:reassignment10} follows one run from the common lower-left start. Initial goals and cells are assigned by $75\,\mathrm{s}$ [Fig.~\ref{fig:s1}]. Because the lower-right region contains no targets, its robot finishes first at $1294\,\mathrm{s}$ [Fig.~\ref{fig:s2}], triggering the reassignment in Fig.~\ref{fig:s3}. A second trigger at $2691\,\mathrm{s}$ produces Figs.~\ref{fig:s4}--\ref{fig:s5}, and all cells are covered by $3174\,\mathrm{s}$ [Fig.~\ref{fig:s6}]. The resulting clustered territories reflect both target density and heterogeneous speed: low-demand cells migrate to robots that would otherwise idle.

\subsection{Evaluation of MDCPP}
Four heterogeneous robots cover an unknown GMM field with unrestricted inter-robot communication. Their travel speeds are $(0.08,0.10,0.30,0.32)\,\mathrm{m/s}$ and minimum service speeds are $(0.015,0.025,0.060,0.070)\,\mathrm{m/s}$. Coverage speed interpolates linearly between these limits according to the cell's normalized Gaussian value.

\subsubsection{Estimation and Quantitative Evaluation}
\begin{figure*}[t]
\centering
\subfloat[Ground truth]{\includegraphics[trim=0 0 0 66,clip,width=0.235\textwidth]{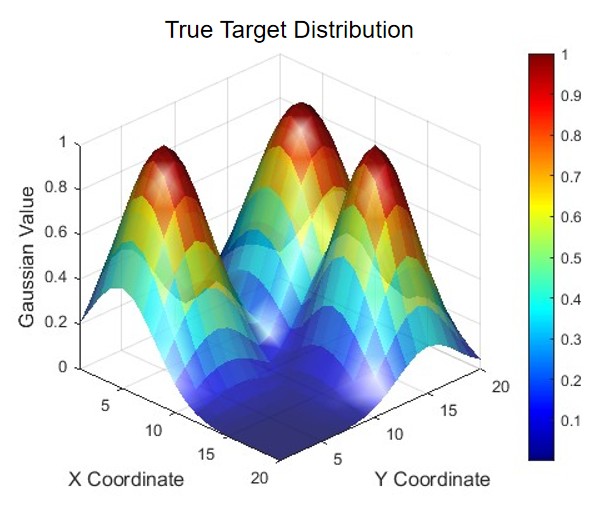}}\hfill
\subfloat[Prediction, 400 s]{\includegraphics[trim=0 0 0 66,clip,width=0.235\textwidth]{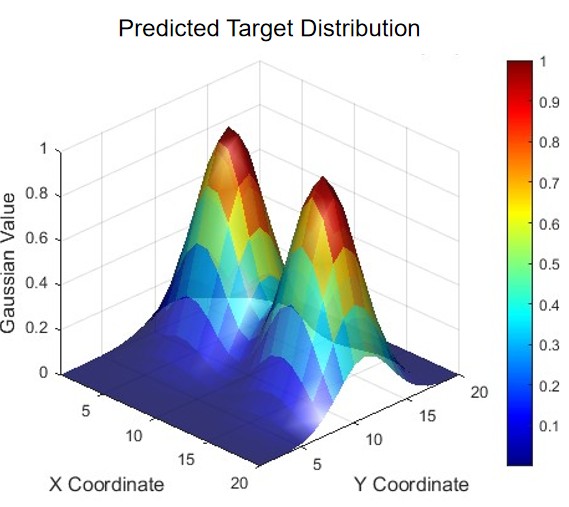}}\hfill
\subfloat[Prediction, 1000 s]{\includegraphics[trim=0 0 0 66,clip,width=0.235\textwidth]{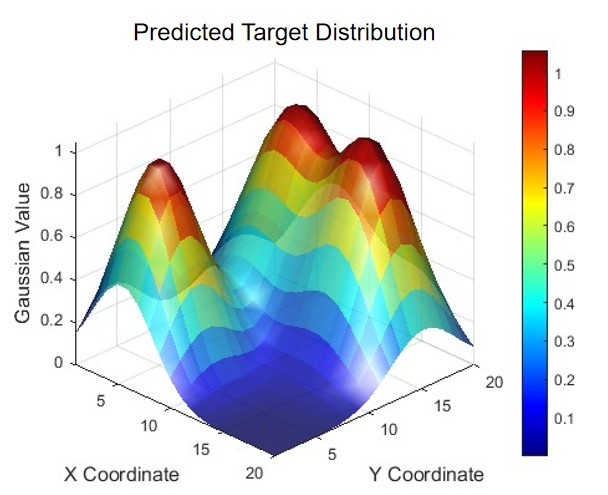}}\hfill
\subfloat[SWD]{\includegraphics[width=0.235\textwidth]{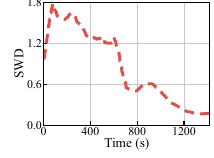}}
\caption{Simulation evaluation: (a)--(c) evolution of workload-field estimation and (d) sliced Wasserstein distance to the ground truth.}
\label{fig:simulation10}
\end{figure*}

The three-hotspot prediction in Fig.~\ref{fig:simulation10}(a)--(c) improves as coverage reveals more cells. Let $p_t$ and $p^\star$ be the unit-mass normalizations of the predicted and true fields. The plotted sliced Wasserstein distance averages one-dimensional Wasserstein distances over $\theta\in\mathbb S^1$,
\begin{equation}
 \operatorname{SWD}(p_t,p^\star)=\mathbb E_{\theta}
 W_1\!\left(\mathcal P_{\theta\#}p_t,\mathcal P_{\theta\#}p^\star\right),
\label{eq:swd}
\end{equation}
where $\mathcal P_{\theta}(x)=\theta^\top x$ and the expectation is approximated by sampled directions \cite{deshpande2018generative}. Figure~\ref{fig:simulation10}(d) declines after the exploratory transient and approaches $0.2$ near $1200\,\mathrm{s}$. The non-monotone early peaks occur because newly discovered high-density cells can temporarily expose a missing component before its center and spread are refitted; the residual error reflects grid discretization and incomplete observations.

\subsection{Baselines and Monte Carlo Comparison}
\begin{figure}[!t]
\centering
\setcounter{subfigure}{4}
\subfloat[Monte Carlo comparison]{\includegraphics[width=0.98\columnwidth]{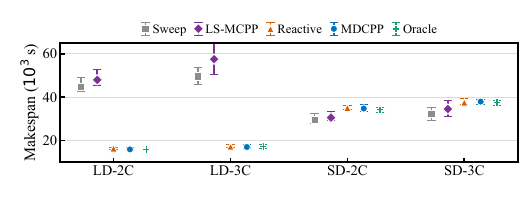}}
\addtocounter{figure}{-1}
\caption{Simulation evaluation (continued): (e) makespan median and IQR over 30 paired instances per scenario and method (600 runs). LD/SD: large/small speed disparity; 2C/3C: two/three workload centers.}
\label{fig:monte_carlo}
\end{figure}

Figure~\ref{fig:monte_carlo}(e) compares fixed rectangular Sweep; geometric LS-MCPP using its official implementation \cite{tang2025lsmcpp}; Reactive, which repartitions from revealed cells without extrapolation; MDCPP; and Oracle, which exposes the true field to the same dynamic allocator. Every policy shares the grid, initial poses, robot speeds, service model, and target field within a paired run. The LD travel/detect pairs are $0.05/0.008$, $0.15/0.030$, $0.30/0.060$, and $0.40/0.080\,\mathrm{m/s}$; SD uses $0.08/0.015$, $0.10/0.020$, $0.12/0.025$, and $0.15/0.030\,\mathrm{m/s}$. We draw 30 instances for each LD/SD and two-/three-center combination by randomizing separated hotspot centers, amplitudes, spreads, and quadrant-stratified starts. LS-MCPP receives only geometry; all routes are evaluated under the realized service times. Dynamic policies use the same trigger rule. Oracle is a full-information reference, not a global lower bound or practical competitor.

Across all 120 paired instances per method, MDCPP reduces median makespan by $40.4\%$ relative to Sweep, $35.3\%$ relative to LS-MCPP, and $1.0\%$ relative to Reactive. Two-sided paired Wilcoxon tests with Holm correction give $p<10^{-6}$, $p<10^{-6}$, and $p=0.026$, respectively. The gain is regime-dependent: relative to Sweep/LS-MCPP, median reductions are $65.3/67.8\%$ in LD--2C and $64.9/70.8\%$ in LD--3C, but MDCPP is $11.0/15.0\%$ slower in SD--2C and $18.8/10.0\%$ slower in SD--3C. Thus prediction compensates for severe service-time heterogeneity, whereas under mild heterogeneity its online-route overhead can outweigh workload balancing. MDCPP's median paired path length is $2.2\%$ shorter than Reactive's, while its makespan is $1.6\%$ higher than Oracle's in median paired relative terms. The archive records raw runs, fields, the LS-MCPP revision, and scripts.

The pooled result should not be interpreted as uniform dominance. MDCPP beats Sweep and LS-MCPP in all 60 LD instances, whereas its pooled win rates over those baselines are $57.5\%$ and $63.3\%$ after the SD cases are included. Against Reactive, it wins $57.5\%$ of all paired instances. These counts agree with the signed-rank analysis: the aggregate advantage is statistically detectable, but prediction should be enabled when estimated service heterogeneity justifies its routing cost rather than treated as an unconditional replacement for geometric allocation.

The crossover is also visible in workload balance and route cost. In LD--2C/3C, MDCPP reduces the median active-time coefficient of variation to $0.163/0.155$, compared with $0.799/0.787$ for Sweep and $0.883/0.915$ for LS-MCPP; the additional transfers are therefore repaid by avoiding a severely overloaded robot. In SD--2C/3C, MDCPP still lowers the coefficient to $0.079/0.067$ from Sweep's $0.255/0.234$, but its median path grows from about $4.29\,$km to $7.63/7.78\,$km. When robot service rates are already similar, this travel increase dominates the smaller balancing benefit. These paired diagnostics explain the regime change rather than attributing it to estimator noise alone.

\subsection{Reproducibility and Computational Cost}
All methods use the same discrete-event executor: a cell earns service credit only on its first visit, while revisits and transfers still add travel. Sweep and LS-MCPP remain fixed. Reactive, MDCPP, and Oracle share the synchronized trigger, nearest-neighbor routing, and serial service-load-decreasing transfers; they differ only in the unobserved-field model (zero prior, fitted mixture, or truth). A deterministic greedy construction starts each epoch, with approach time only as a regularizer; transfers then decrease service-time makespan. Thus the benchmark instantiates the service component of \eqref{eq:workload}, not exact joint route optimization. The archive records all seeds and fields and pins LS-MCPP to revision bb06f1d with 2000 local-search iterations. LS-MCPP is therefore a reproducible geometric baseline, while Oracle is a full-information reference for the implemented heuristic, not a global optimum.

The benchmark uses a $20\times20$ grid with $10\,$m cells and four robots. A deterministic hash fixes each scenario-replicate pair. Hotspot centers are separated by at least five cells; amplitudes are drawn uniformly from $[0.75,1]$, spreads from $[2.2,3.8]$ cells, and the summed field is clipped to unit density. Starts are stratified one per quadrant and robot identities are shuffled, so all methods receive identical paired inputs. Cell observations are noise-free in this benchmark to isolate the value of workload extrapolation; robustness to bounded observation-induced cost error is addressed by Proposition~2. Dynamic policies reassess after the fastest active route completes $35\%$ of its queue, controlling update frequency independently of the estimator.

MDCPP uses the stated constant fallback until the sample prerequisites hold, then fits one to three nonnegative Gaussian components and overwrites predictions at measured cells. Tests pair methods by scenario and replicate; gains are medians of 120 paired relative differences. Holm correction is applied jointly to the four comparator tests, preventing field difficulty and start placement from becoming between-method variation.

Median recorded computation times are $3.235\,$s for LS-MCPP, $0.095\,$s for Reactive, $0.176\,$s for MDCPP, and $0.110\,$s for Oracle; Sweep is negligible. LS-MCPP records solver time, whereas each dynamic value covers its event loop, so these groups are not directly comparable. Within the common dynamic loop, MDCPP adds $0.081\,$s over Reactive per mission. These serial-Python workstation measurements exclude startup and are not embedded-controller latencies. Pooled median travel distances are $4.29$, $4.12$, $7.56$, $7.41$, and $7.20\,$km for Sweep, LS-MCPP, Reactive, MDCPP, and Oracle. Shorter geometric routes but longer LD completion times confirm that the objective is service-weighted makespan; the opposite SD result identifies when extra routing is not worthwhile.

Oracle occasionally loses to MDCPP because both use a pairwise-local allocator and nearest-neighbor routing: exact costs can induce a workload-reducing transfer whose added route distance is unfavorable after heuristic ordering, while prediction error can fortuitously avoid it. Oracle therefore measures perfect workload information within the implemented stack; it is neither a global optimum nor a per-instance lower bound.

\section{Hardware Experiments}
\subsection{Platform and Protocol}
We evaluated the simulation-to-real transfer on three Wheeltec omnidirectional UGVs ($0.30\,\mathrm{m}\times0.24\,\mathrm{m}$), each equipped with a Raspberry Pi 4B, wheel odometry, and a planar lidar. The robots ran Ubuntu 22.04 and ROS~2 Humble and were localized by AMCL in a prebuilt map. A MATLAB R2024a controller communicated over Wi-Fi using an independent ROS~2 domain for each robot. The controller directly published body-frame mecanum velocity commands; no Nav2 local planner was used.

The simulation grid was scaled to a $4.5\,\mathrm{m}\times4.5\,\mathrm{m}$ arena. Planned curves were sampled into collision-free waypoint sequences with a nominal $0.30\,\mathrm{m}$ spacing and a $0.20\,\mathrm{m}$ arrival tolerance. After a separate staging phase, the logged coverage phase executed 12, 11, and 15 waypoints for UGV1--UGV3. At position $q_i$, the commanded speed magnitude was
\begin{equation}
v_i(q_i)=v_{\mathrm{tr},i}-(v_{\mathrm{tr},i}-v_{\mathrm{det},i})
\max_k\exp\!\left(-\frac{\|q_i-\mu_k\|^2}{2\sigma_k^2}\right),
\label{eq:hardware_speed}
\end{equation}
clipped to $[v_{\mathrm{det},i},v_{\mathrm{tr},i}]$. The travel/detection pairs were $(0.080,0.015)$, $(0.100,0.025)$, and $(0.150,0.060)\,\mathrm{m/s}$, and the two physical hotspots shared $\sigma=1.129\,\mathrm{m}$.
For command generation, the maximum component activation is used instead of the clipped sum in \eqref{eq:rho_true}; this keeps the commanded speed within the calibrated interval without an additional overlap normalization.

\subsection{Metrics and Data Processing}
Let $q_j$ be an AMCL sample and $[a_m,b_m]$ a nonzero reference segment $m$. With $u_m=b_m-a_m$, cross-track error is
\begin{align}
 e_\perp(q_j)&=\min_m\|q_j-a_m-\lambda_{jm}u_m\|,\label{eq:cross_track_final}\\
 \lambda_{jm}&=\operatorname{clip}_{[0,1]}
 \frac{(q_j-a_m)^\top u_m}{\|u_m\|^2}.
\end{align}
Speed is the norm of planar odometry velocity after a second-order $0.5\,\mathrm{Hz}$ low-pass filter. We report speed error against \eqref{eq:hardware_speed} over the complete active interval and over steady segments excluding $\pm1.6\,\mathrm{s}$ around waypoint changes. For steady segments, a causal lag up to $2\,\mathrm{s}$ is selected by minimum RMSE. The same cutoff, exclusion window, and lag search are used for all robots. Because the data contain one synchronized run, time samples are not treated as independent replicates and no significance test is reported.

The common log contains 620 synchronized pose/odometry samples at a median $4.94\,\mathrm{Hz}$. Polyline projection keeps tracking error well-defined between waypoints. Speed RMSE measures command-following accuracy, whereas $R^2$ tests whether measured variation follows the spatial command shape and is meaningful only with sufficient command variance. After transition exclusion, the steady masks retain 173, 288, and 380 samples for UGV1--UGV3; the selected response lags are 0.810, 0.405, and $0.607\,\mathrm{s}$.

Low-pass filtering suppresses encoder-scale fluctuations, the common exclusion window removes acceleration and turning intervals without robot-specific tuning, and causal lag alignment accounts for drivetrain response. Full-interval errors are retained so the steady-state fit is not presented as total closed-loop performance.

\begin{table}[t]
\centering
\small
\setlength{\tabcolsep}{3.2pt}
\caption{Three-UGV Hardware Results}
\label{tab:hardware_results}
\begin{tabular}{lrrrr}
\hline
 & WPs & $T_i$ (s) & Track RMSE (m) & Speed RMSE/$R^2$\\
\hline
UGV1 & 12/12 & 71.1  & 0.033 & 0.0036/0.060\\
UGV2 & 11/11 & 91.8  & 0.040 & 0.0037/0.976\\
UGV3 & 15/15 & 125.4 & 0.035 & 0.0052/0.936\\
\hline
\end{tabular}
\end{table}

\begin{figure}[!t]
\centering
\subfloat[Three-UGV trajectories]{\includegraphics[width=0.50\columnwidth]{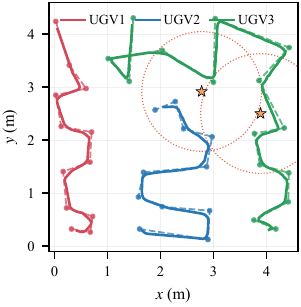}}\\[-0.8mm]
\subfloat[Speed command validation]{\includegraphics[width=0.68\columnwidth]{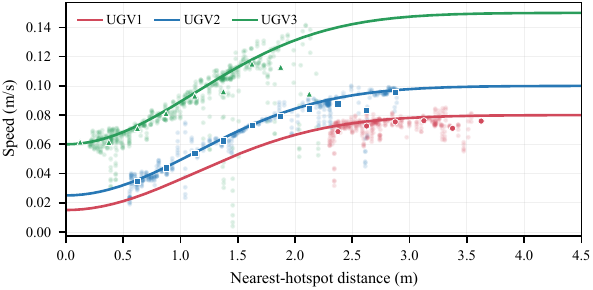}}
\caption{Hardware validation. (a) Reference polylines (dashed), AMCL trajectories (solid), hotspot centers (stars), and one-standard-deviation contours (dotted). (b) Filtered speed samples (faint points), bin means (markers), and command curves (solid).}
\label{fig:hardware_validation}
\end{figure}

\subsection{Results and Discussion}
Figure~\ref{fig:hardware_validation}(a) and Table~\ref{tab:hardware_results} summarize route execution. UGV1--UGV3 reached all 38 waypoints in 71.1, 91.8, and $125.4\,\mathrm{s}$. Their cross-track RMSEs were 0.033, 0.040, and $0.035\,\mathrm{m}$; MAEs were 0.027, 0.033, and $0.028\,\mathrm{m}$; and 95th-percentile errors were 0.061, 0.076, and $0.074\,\mathrm{m}$, all below $1.7\%$ of the arena width. Measured path lengths were 5.13, 6.05, and $10.58\,\mathrm{m}$.

In Fig.~\ref{fig:hardware_validation}(b), UGV2 and UGV3 traverse informative distance ranges and yield steady speed RMSEs of 0.0037 and $0.0052\,\mathrm{m/s}$, with $R^2=0.976$ and $0.936$; their MAEs are 0.0025 and $0.0032\,\mathrm{m/s}$. UGV1 remains in the saturated far field ($2.30$--$3.53\,\mathrm{m}$), so its $0.0036\,\mathrm{m/s}$ RMSE verifies tracking but $R^2=0.060$ is uninformative. Full-interval RMSEs rise to 0.0074, 0.0077, and $0.0177\,\mathrm{m/s}$ because turns and waypoint switches introduce dynamics outside the steady-state spatial law.

The hardware run separates two interfaces: cross-track error tests waypoint execution and speed error tests spatial modulation. It does not test online GMM fitting or physical reassignment, which remain simulation claims. Because the waypoints are precomputed, the experiment cannot establish end-to-end policy superiority. The localized, slowly changing GMM assumption also excludes diffuse or moving fields, for which the estimator and reassignment schedule would require modification.

\subsection{Cross-domain Interpretation}
Simulation isolates the allocation benefit of predicting unobserved demand; hardware tests whether the resulting waypoint and speed commands remain executable under localization noise, drivetrain lag, and wireless control. Cross-track RMSE is below $0.9\%$ of the arena width and steady speed RMSE is $3.5\%$--$4.6\%$ of travel speed. Thus the simulated allocator is paired with a physically realizable command profile without conflating interface validation with an unperformed end-to-end comparison.

Because arena scale, speeds, and routes differ, cross-domain completion times are not compared; the transferable object is the normalized density-to-speed law. The spatially separated hardware routes test this interface, not the Monte Carlo effect size or inter-robot collision avoidance.

\section{Conclusions}
This paper presented MDCPP, which learns spatial service demand from partial observations and repartitions uncovered cells by predicted remaining completion time. Under the stated assumptions, each assignment round terminates at a pairwise local optimum, bounded estimation error yields a bounded service-time makespan loss, and the resulting routes achieve complete coverage. Monte Carlo comparisons show that prediction substantially improves makespan under strong heterogeneity and modestly improves the aggregate reactive comparison, while mild heterogeneity exposes its route-overhead limit. A three-UGV experiment confirms executable waypoint tracking and Gaussian speed modulation. These results establish the value and scope of workload-aware dynamic coverage allocation.

\bibliographystyle{IEEEtran}
\bibliography{reference_ral_final}

@article{galceran2013survey,
  title={A survey on coverage path planning for robotics},
  author={Galceran, Enric and Carreras, Marc},
  journal={Robotics and Autonomous Systems},
  volume={61},
  number={12},
  pages={1258--1276},
  year={2013},
  publisher={Elsevier}
}

@article{oksanen2009coverage,
  title={Coverage path planning algorithms for agricultural field machines},
  author={Oksanen, Timo and Visala, Arto},
  journal={Journal of Field Robotics},
  volume={26},
  number={8},
  pages={651--668},
  year={2009},
  publisher={Wiley Online Library}
}

@article{cabreira2018energy,
  title={Energy-aware spiral coverage path planning for {UAV} photogrammetric applications},
  author={Cabreira, Taua M and Di Franco, Carmelo and Ferreira, Paulo R and Buttazzo, Giorgio C},
  journal={IEEE Robotics and Automation Letters},
  volume={3},
  number={4},
  pages={3662--3668},
  year={2018},
  publisher={IEEE}
}

@article{otto2018optimization,
  title={Optimization approaches for civil applications of unmanned aerial vehicles (UAVs) or aerial drones: A survey},
  author={Otto, Alena and Agatz, Niels and Campbell, James and Golden, Bruce and Pesch, Erwin},
  journal={Networks},
  volume={72},
  number={4},
  pages={411--458},
  year={2018},
  publisher={Wiley Online Library}
}

@article{acar2002sensor,
  title={Sensor-based coverage of unknown environments: Incremental construction of {Morse} decompositions},
  author={Acar, Ercan U and Choset, Howie},
  journal={The International Journal of Robotics Research},
  volume={21},
  number={4},
  pages={345--366},
  year={2002},
  publisher={SAGE Publications Sage UK: London, England}
}

@inproceedings{karapetyan2017efficient,
  title={Efficient multi-robot coverage of a known environment},
  author={Karapetyan, Nare and Benson, Kelly and McKinney, Chris and Taslakian, Perouz and Rekleitis, Ioannis},
  booktitle={2017 IEEE/RSJ International Conference on Intelligent Robots and Systems (IROS)},
  pages={1846--1852},
  year={2017},
  organization={IEEE}
}

@inproceedings{fazli2010complete,
  title={Complete and robust cooperative robot area coverage with limited range},
  author={Fazli, Pooyan and Davoodi, Alireza and Pasquier, Philippe and Mackworth, Alan K},
  booktitle={Proc. IEEE/RSJ Int. Conf. Intelligent Robots and Systems (IROS)},
  pages={5577--5582},
  year={2010}
}

@article{rekleitis2008efficient,
  title={Efficient boustrophedon multi-robot coverage: an algorithmic approach},
  author={Rekleitis, Ioannis and New, Ai Peng and Rankin, Edward Samuel and Choset, Howie},
  journal={Annals of Mathematics and Artificial Intelligence},
  volume={52},
  number={2},
  pages={109--142},
  year={2008},
  publisher={Springer}
}

@inproceedings{tang2021mstc,
  title={{MSTC}$\ast$: Multi-Robot Coverage Path Planning Under Physical Constraints},
  author={Tang, Jingtao and Sun, Chun and Zhang, Xinyu},
  booktitle={2021 IEEE International Conference on Robotics and Automation (ICRA)},
  pages={2518--2524},
  year={2021},
  organization={IEEE}
}

@inproceedings{collins2021scalable,
  title={Scalable coverage path planning of multi-robot teams for monitoring non-convex areas},
  author={Collins, Leighton and Ghassemi, Payam and Esfahani, Ehsan T and Doermann, David and Dantu, Karthik and Chowdhury, Souma},
  booktitle={2021 IEEE International Conference on Robotics and Automation (ICRA)},
  pages={7393--7399},
  year={2021},
  organization={IEEE}
}

@inproceedings{tolstaya2021multi,
  title={Multi-robot coverage and exploration using spatial graph neural networks},
  author={Tolstaya, Ekaterina and Paulos, James and Kumar, Vijay and Ribeiro, Alejandro},
  booktitle={2021 IEEE/RSJ International Conference on Intelligent Robots and Systems (IROS)},
  pages={8944--8950},
  year={2021},
  organization={IEEE}
}

@article{zhu2019complete,
  title={Complete coverage path planning of autonomous underwater vehicle based on GBNN algorithm},
  author={Zhu, Daqi and Tian, Chen and Sun, Bing and Luo, Chaomin},
  journal={Journal of Intelligent \& Robotic Systems},
  volume={94},
  number={1},
  pages={237--249},
  year={2019},
  publisher={Springer}
}

@article{balzer2009capacity,
  title={Capacity-constrained point distributions: A variant of Lloyd's method},
  author={Balzer, Michael and Schl{\"o}mer, Thomas and Deussen, Oliver},
  journal={ACM Transactions on Graphics (TOG)},
  volume={28},
  number={3},
  pages={1--8},
  year={2009},
  publisher={ACM New York, NY, USA}
}

@inproceedings{chen2021distributed,
  title={Distributed multi-target tracking for heterogeneous mobile sensing networks with limited fields of view},
  author={Chen, Jun and Dames, Philip},
  booktitle={2021 IEEE International Conference on Robotics and Automation (ICRA)},
  pages={9058--9064},
  year={2021},
  organization={IEEE}
}

@article{cortes2004coverage,
  title={Coverage control for mobile sensing networks},
  author={Cortes, Jorge and Martinez, Sonia and Karatas, Timur and Bullo, Francesco},
  journal={IEEE Transactions on Robotics and Automation},
  volume={20},
  number={2},
  pages={243--255},
  year={2004},
  publisher={IEEE}
}

@book{applegate2011traveling,
  title={The Traveling Salesman Problem: A Computational Study},
  author={Applegate, David L and Bixby, Robert E and Chv{\'a}tal, Va{\v{s}}ek and Cook, William J},
  year={2007},
  publisher={Princeton University Press},
  address={Princeton, NJ, USA}
}

@article{hu2023multi,
  title={Multi-UAV coverage path planning: A distributed online cooperation method},
  author={Hu, Wenjian and Yu, Yao and Liu, Shumei and She, Changyang and Guo, Lei and Vucetic, Branka and Li, Yonghui},
  journal={IEEE Transactions on Vehicular Technology},
  volume={72},
  number={9},
  pages={11727--11740},
  year={2023},
  publisher={IEEE}
}

@article{chen2024improved,
  title={Improved coverage path planning for indoor robots based on BIM and robotic configurations},
  author={Chen, Zhengyi and Wang, Hao and Chen, Keyu and Song, Changhao and Zhang, Xiao and Wang, Boyu and Cheng, Jack CP},
  journal={Automation in Construction},
  volume={158},
  pages={105160},
  year={2024},
  publisher={Elsevier}
}

@inproceedings{deshpande2018generative,
  title={Generative modeling using the sliced {Wasserstein} distance},
  author={Deshpande, Ishan and Zhang, Ziyu and Schwing, Alexander G},
  booktitle={Proc. IEEE Conf. Computer Vision and Pattern Recognition (CVPR)},
  pages={3483--3491},
  year={2018}
}

@article{crawford2020use,
  title={The use of Gaussian mixture models with atmospheric Lagrangian particle dispersion models for density estimation and feature identification},
  author={Crawford, Alice},
  journal={Atmosphere},
  volume={11},
  number={12},
  pages={1369},
  year={2020},
  publisher={MDPI}
}

@article{tang2023mip,
  author  = {Jingtao Tang and Hang Ma},
  title   = {Mixed Integer Programming for Time-Optimal Multi-Robot Coverage Path Planning with Efficient Heuristics},
  journal = {IEEE Robotics and Automation Letters},
  volume  = {8}, number = {10}, pages = {6491--6498}, year = {2023},
  doi     = {10.1109/LRA.2023.3306996}
}

@article{tang2025lsmcpp,
  author  = {Jingtao Tang and Zining Mao and Hang Ma},
  title   = {Large-Scale Multirobot Coverage Path Planning on Grids With Path Deconfliction},
  journal = {IEEE Transactions on Robotics},
  volume  = {41}, pages = {3348--3367}, year = {2025},
  doi     = {10.1109/TRO.2025.3567476}
}

@article{wang2024apf,
  author  = {Zikai Wang and Xin Zhao and Jiekai Zhang and Ning Yang and Pengyu Wang and Jingtao Tang and Jianan Zhang and Ling Shi},
  title   = {{APF-CPP}: An Artificial Potential Field Based Multi-Robot Online Coverage Path Planning Approach},
  journal = {IEEE Robotics and Automation Letters},
  volume  = {9}, number = {11}, pages = {9199--9206}, year = {2024},
  doi     = {10.1109/LRA.2024.3432351}
}

@article{wang2025mac,
  author  = {Zikai Wang and Xiaoxu Lyu and Jiekai Zhang and Pengyu Wang and Yuxing Zhong and Ling Shi},
  title   = {{MAC-Planner}: A Novel Task Allocation and Path Planning Framework for Multi-Robot Online Coverage Processes},
  journal = {IEEE Robotics and Automation Letters},
  volume  = {10}, number = {5}, pages = {4404--4411}, year = {2025},
  doi     = {10.1109/LRA.2025.3551078}
}

@inproceedings{mo2024mdstc,
  author    = {Weimin Mo and Zhiyun Lin},
  title     = {{MDSTC}: A Dynamic Approach to Multi-Robot Coverage Path Planning},
  booktitle = {Proc. Int. Conf. Control, Automation, Robotics and Vision (ICARCV)},
  pages     = {1100--1105}, year = {2024},
  doi       = {10.1109/ICARCV63323.2024.10821576}
}

@inproceedings{mitra2025concurrent,
  author    = {Ratijit Mitra and Indranil Saha},
  title     = {Online Concurrent Multi-Robot Coverage Path Planning},
  booktitle = {Proc. IEEE/RSJ Int. Conf. Intelligent Robots and Systems (IROS)},
  pages     = {8691--8698}, year = {2025},
  doi       = {10.1109/IROS60139.2025.11247261}
}

@inproceedings{mitra2024ondemand,
  author    = {Ratijit Mitra and Indranil Saha},
  title     = {Online On-Demand Multi-Robot Coverage Path Planning},
  booktitle = {Proc. IEEE Int. Conf. Robotics and Automation (ICRA)},
  pages     = {14583--14589}, year = {2024},
  doi       = {10.1109/ICRA57147.2024.10611610}
}

@article{luo2026reconfigurable,
  author  = {Cai Xia Luo and Jintao Guo and Zixuan Liu and Lei Liu and Chunbo Luo},
  title   = {Real-Time Path-Reconfigurable Coverage Planning for Multi-{UAV} Missions Over Disjoint Areas},
  journal = {IEEE Robotics and Automation Letters},
  volume  = {11}, number = {2}, pages = {1226--1233}, year = {2026},
  doi     = {10.1109/LRA.2025.3641039}
}

@article{kim2026graph,
  author  = {Kyungseo Kim and Jinwhan Kim},
  title   = {Efficient Graph-Based Area Partitioning for Balanced Multi-Robot Coverage Path Planning},
  journal = {Robotics and Autonomous Systems},
  volume  = {201}, pages = {105445}, year = {2026},
  doi     = {10.1016/j.robot.2026.105445}
}
\end{document}